\documentclass[11pt]{article}

\usepackage[preprint]{acl}
\usepackage{amssymb}
\usepackage{times}
\usepackage{latexsym}

\usepackage[T1]{fontenc}

\usepackage[utf8]{inputenc}

\usepackage{microtype}

\usepackage{inconsolata}
\usepackage{booktabs}
\usepackage{graphicx}
\usepackage{multirow}
\usepackage{amsmath}
\usepackage{xcolor}
\usepackage{colortbl}
\usepackage{multirow} 
\usepackage{tabularx}
\usepackage[inline]{enumitem}
\usepackage{tabularx} 
\usepackage{booktabs}
\usepackage[most]{tcolorbox}
\newtcolorbox{promptbox}[1][]{
    colback=gray!10!white,    
    colframe=black!75,        
    boxrule=0.6pt,            
    arc=4pt,                  
    left=8pt,                 
    right=8pt,                
    top=8pt,                  
    bottom=8pt,               
    enhanced,                 
    fontupper=\small\itshape, 
    #1                        
}

\newcolumntype{g}{>{\columncolor{gray!10}}c}

\title{Decoupled Reasoning with Implicit Fact Tokens (DRIFT): A Dual-Model Framework for Efficient Long-Context Inference
}

\author{
  Wenxuan Xie$^{1,2,3}$, 
  Yujia Wang$^{4}$, 
  Xin Tan$^{1}$, 
  Chaochao Lu$^{1}$, 
  Xia Hu$^{1}$, 
  \textbf{Xuhong Wang}$^{1,}$\thanks{~~Corresponding author.} \\
  $^1$Shanghai Artificial Intelligence Laboratory, Shanghai, China \\
  $^2$Fudan University, Shanghai, China \\
  $^3$Shanghai Innovation Institute, Shanghai, China \\
  $^4$Tongji University, Shanghai, China \\
  \texttt{wxxie25@m.fudan.edu.cn}, \texttt{wangxuhong@pjlab.org.cn}
} 

\begin{document}
\maketitle
\begin{abstract}
The integration of extensive, dynamic knowledge into Large Language Models (LLMs) remains a significant challenge due to the inherent entanglement of factual data and reasoning patterns. Existing solutions, ranging from non-parametric Retrieval-Augmented Generation (RAG) to parametric knowledge editing, are often constrained in practice by finite context windows, retriever noise, or the risk of catastrophic forgetting. In this paper, we propose \textbf{DRIFT}, a novel dual-model architecture designed to explicitly decouple knowledge extraction from the reasoning process. Unlike static prompt compression, DRIFT employs a lightweight knowledge model to dynamically compress document chunks into implicit fact tokens conditioned on the query. These dense representations are projected into the reasoning model's embedding space, replacing raw, redundant text while maintaining inference accuracy. Extensive experiments show that DRIFT significantly improves performance on \textbf{long-context tasks}, outperforming strong baselines among comparably sized models. Our approach provides a scalable and efficient paradigm for extending the effective context window and reasoning capabilities of LLMs. Our code is available at \url{https://github.com/Lancelot-Xie/DRIFT}.
\end{abstract}

\section{Introduction}

The applicability of Large Language Models (LLMs) to knowledge-intensive tasks is limited by the static nature of their pre-training data. To address this limitation, prior work has explored two complementary strategies.The first focuses on augmenting the input context, while the second emphasizes knowledge parameter editing.

Traditional input augmentation via RAG or long-context prompting is increasingly constrained by the ``retriever's ceiling'' and the quadratic computational costs of processing long sequences. Neural compression methods (e.g., COCOM \citep{cocom}, C3 \citep{c3}) attempt to mitigate this by distilling text into latent representations; however, as they primarily focus on static compression, task-critical information relevant to the query is frequently lost. Conversely, internalizing knowledge through direct parametric updates, such as fine-tuning or knowledge editing, often disrupts the inherent coupling between a model's internal knowledge and its reasoning logic, while also risking catastrophic forgetting. While modular approaches like MLP Memory \citep{MLP_memory} offer a plug-and-play alternative, they remain bound to pre-indexed resources and struggle to handle instantaneous, unseen long-context inputs in real-time.

To address these challenges, we introduce \textbf{DRIFT}, a dual-model architecture that decouples context processing from core reasoning. In this framework, a lightweight \textbf{knowledge model} extracts query-relevant information from document chunks and compresses it into high-density \textbf{implicit fact tokens} within a latent space. These tokens serve as concise, knowledge-rich representations that are projected into a larger \textbf{reasoning model's} embedding space. The reasoning model can perform sophisticated inference efficiently based on this compact factual context instead of raw text, even in long-context or knowledge-intensive scenarios. By delegating the processing of redundant background knowledge to the knowledge module, this design allows the reasoning model to remain unburdened by raw context, focusing instead on "clean" and deep inference through the distilled factual representations.

Our main contributions are as follows:
\begin{itemize}
    \item \textbf{A Decoupled Inference Paradigm for Large Language Models:} 
We propose a dual-model framework that decouples knowledge extraction from reasoning. A lightweight knowledge model encodes query-relevant information into compact fact tokens, which are consumed by a larger reasoning model for inference. Compared with directly processing long contexts, DRIFT improves performance while substantially reducing inference latency, achieving \textbf{an average 7× speedup on 256k-token documents}.
    
    \item \textbf{Expanding Effective Context Window with High-Ratio Compression:} 
    By encoding extensive textual knowledge into compact fact tokens, our framework significantly extends the model's usable context window. 
    Specifically, our \textbf{DRIFT} model (based on Mistral 7B) achieves a \textbf{32$\times$ compression ratio} while \textbf{improving accuracy from 20.87\% to 29.22\%} on the LongBench v2 benchmark, demonstrating superior reasoning capabilities with substantially reduced time overhead. Furthermore, even under more aggressive compression settings (\textbf{64}$\times$ and \textbf{128}$\times$), DRIFT remains highly competitive, indicating strong robustness to extreme compression ratios. 
    
    \item \textbf{Comprehensive Empirical Analysis and Resource Contribution:} 
 We construct a large-scale Document–QA–Evidence dataset with fine-grained supervision, comprising over \textbf{300K} instances and documents ranging from \textbf{1K} to \textbf{8K} tokens. We further conduct extensive ablation studies and quantitative evaluations, rigorously validating the framework and demonstrating its robustness.
\end{itemize}

\section{Related Work}

\subsection{Prompt Compression}
Directly feeding new knowledge as context into Large Language Models (LLMs) is constrained by limited context windows and incurs significant overhead in both memory and computation. To mitigate these issues, various prompt compression methods have been proposed, which generally fall into two paradigms: \textit{Hard Compression} and \textit{Soft Compression}.

\paragraph{Hard Compression (Token Selection).}
Hard compression methods, also known as token pruning or selection, aim to reduce input length by discarding tokens deemed less informative. Early approaches like Selective Context \citep{selective_context} utilize self-information or perplexity metrics to filter out redundancy. More advanced frameworks, such as LongLLMLingua \citep{longllmlingua} and LLMLingua-2 \citep{llmlingua2}, perform prompt compression via task-aware coarse-to-fine filtering and distillation-based token selection, respectively.
Despite these advancements, hard compression approaches inherently limit the model's reasoning potential. They \textbf{irreversibly discard information} and rely on rigid, locally made retention decisions that often fail to preserve the \textbf{global semantic structure} required for reliable complex reasoning.

\paragraph{Soft Compression (Latent Representation).}
Early soft compression approaches, such as AutoCompressor \citep{autocompressor}, Gist Tokens \citep{gist_tokens}, and ICAE \citep{icae}, integrate compression and reasoning within a single language model. While effective for moderate context reduction, this tightly coupled design makes it difficult for a single model to simultaneously excel at both high-fidelity compression and complex reasoning, particularly under extreme compression ratios. To address this limitation, subsequent work explores decoupling compression from reasoning. Methods such as xRAG \citep{xrag} and COCOM \citep{cocom} build upon the Retrieval-Augmented Generation (RAG) paradigm by compressing retrieved documents into compact latent representations before passing them to the language model. Although effective in reducing input length, these approaches remain fundamentally constrained by the retrieval stage and inherit the upper-bound limitations of RAG systems. Beyond RAG-based compression, E2LLM \citep{E2LLM} employs a lightweight encoder to compress long inputs into latent representations for downstream LLM reasoning, though the model weights are not publicly released, limiting reproducibility and practical adoption. Context Cascade Compression (C3) \citep{c3} demonstrates strong compression fidelity but lacks task-specific adaptation for downstream reasoning. As a result, a gap remains between compressed representation understanding and effective reasoning.

However, most existing soft compression methods operate in a \textbf{static} and query-agnostic manner, producing generic compressed representations that often fail to preserve task-critical information under high compression ratios. In contrast, \textbf{DRIFT} adopts a \textbf{dynamic, query-conditioned compression strategy} that selectively encodes relevant information, enabling effective reasoning even under \textbf{extreme compression ratios} in knowledge-intensive scenarios.

\subsection{Learned Memory}
nother line of work introduces learned parametric memory modules, such as Memory Decoder \citep{memory_decoder} and MLP Memory \citep{MLP_memory}, which store knowledge in trainable parameters and are often pretrained to emulate retrieval behavior. These methods reduce inference latency without modifying the underlying model parameters, thereby preserving general reasoning capabilities. However, these architectures are inherently static-resource bound: they rely on pre-indexing or offline training on fixed knowledge bases, rendering them incapable of handling instantaneous, long-context inputs in real-time. Moreover, as these modules are often pre-trained to emulate or compress retriever behaviors, their inference dynamics remain tethered to the limitations of the original retrieval paradigm.

\section{Methodology}

Our proposed strategy encourages the knowledge model to perform information-adaptive compression, rather than static compression. This compels the model to first identify and abstract core informational content before encoding it into a compressed semantic representation. As a result, the model learns to prioritize semantically meaningful content over superficial token patterns, leading to improved generalization and robustness in downstream tasks. 

\subsection{Bucketed Compression: Beyond Fixed-Ratio Compression}
Most existing context compression methods adopt a fixed-ratio strategy, where the number of compressed tokens is strictly proportional to the input length (e.g., compressing 128 input tokens into 16 output tokens for an 8:1 ratio). However, such a design implicitly assumes that informative content is evenly distributed across the input, which rarely holds in real-world tasks. The amount of query-relevant information within a given context is always unknown and varies in token length. Certain tokens (such as numbers, named entities, main verbs, and constraint-related words) carry a large amount of task-relevant information, whereas redundant modifiers and generic, content-free sentences contribute little to understanding the context. For example, in document-grounded QA or long-form reasoning, a few critical sentences may carry the majority of the answer-relevant information.

This fixed-ratio compression can thus become fragile in scenarios where the input contains sparse but crucial evidence. Moreover, training a model to uniformly compress variable-length sequences may encourage shortcut learning (e.g., positional bias, over-averaging), hindering semantic abstraction. To address this, we propose a Bucketed Compression strategy that shifts from ratio-based compression to range-based compression. Instead of computing the number of output tokens as a fixed fraction of the input, we predefine token-length buckets (e.g., 64–128, 128–256), and map each input in a bucket to a fixed-size output based on the upper bound of that range.

To make the distinction clear, we present a direct comparison of the two strategies in formulaic form, under the assumption of a compression ratio $c$:
\[
\xi_{\text{uniform}} = \left\lceil \tfrac{n}{c} \right\rceil, 
\qquad
\xi_{\text{bucket}} = \left\lceil \tfrac{b(n)}{c} \right\rceil 
\]
where $b(n)$ denotes the upper bound of the bucket containing $n$ tokens.

\subsection{DRIFT: Decoupled Reasoning with Implicit Fact Tokens}
The core idea of DRIFT is to modularize and explicitly separate knowledge reading and reasoning. Specifically, a small-scale knowledge model ($\psi_{kno}$) is responsible for reading long documents and compressing them into query-relevant information, while a large-scale reasoning model {$\psi_{kno}$} focuses on utilizing this compressed knowledge to perform complex reasoning and generate answers. The interaction between the two models is realized in the latent space, which reduces redundancy and mitigates the risk of irrelevant noise interfering with the reasoning process.

\begin{figure*}[t]
    \centering
    \includegraphics[width=1\linewidth]{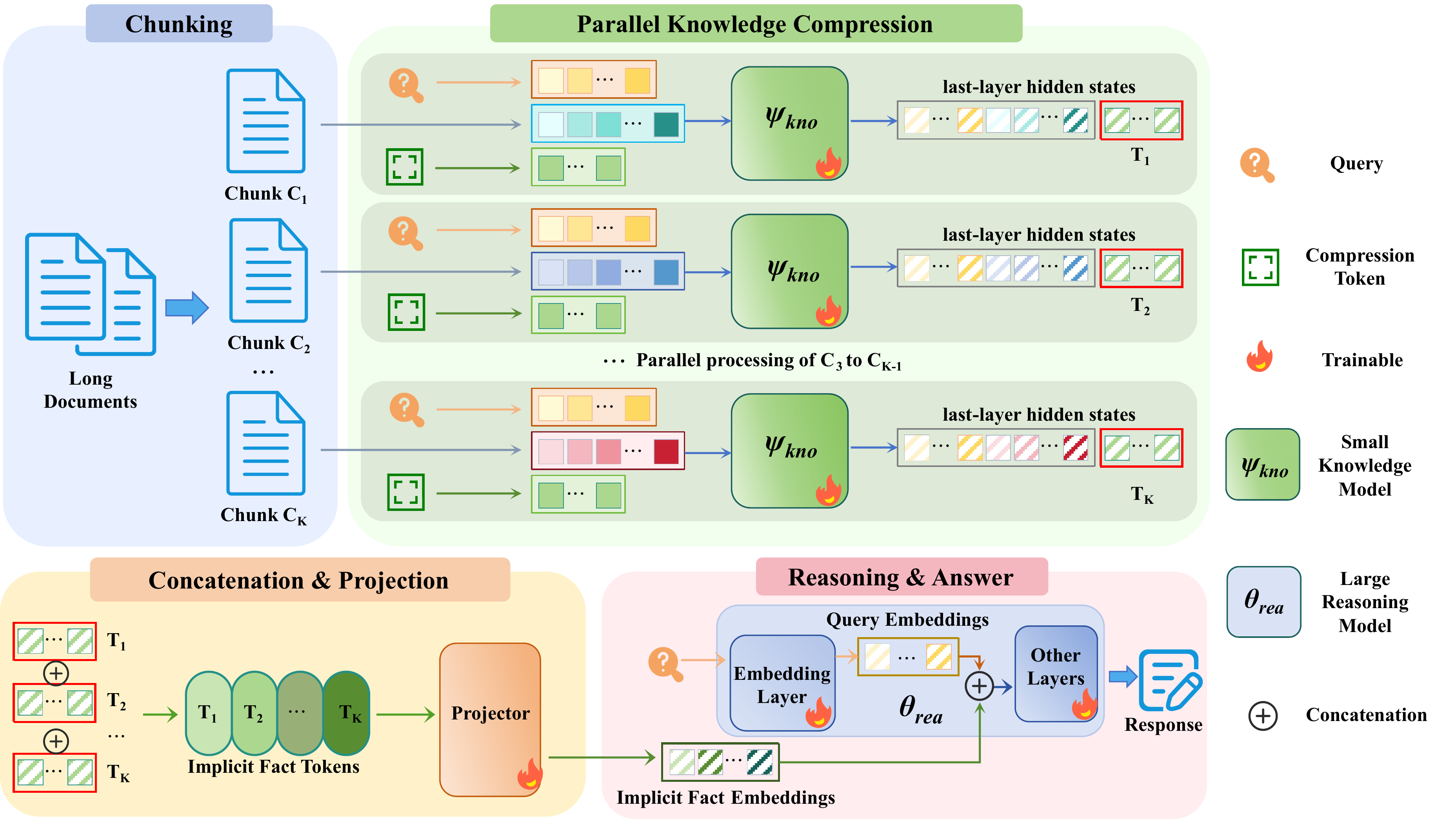}
    \caption{\textbf{The overall workflow of DRIFT.} DRIFT implements knowledge compression and decoupled reasoning in four steps.
    \textbf{Step 1:} The long document $X$ is recursively partitioned into semantically coherent chunks to preserve structural integrity.
    \textbf{Step 2: } The small knowledge model $\psi_{kno}$ compresses query-relevant information from each chunk in parallel into latent implicit fact tokens $T_J$.
    \textbf{Step 3: } The latent tokens are concatenated and mapped by an MLP projector $\pi$ to align with the reasoning model's embedding space. 
    \textbf{Step 4: } The large reasoning model $\theta_{rea}$ generates the final response by performing efficient inference on the concatenated embeddings.}
    \label{fig:DRIFT}
\end{figure*}

To facilitate a clear understanding of our method, the overall workflow of DRIFT is depicted in Figure \ref{fig:DRIFT}. We first define a document as $X = (x_1, \dots, x_n)$, where $n$ is the number of tokens in the document.  For long input contexts, a systematic document chunking strategy was employed within the DRIFT framework. We utilized the \texttt{RecursiveCharacterTextSplitter} from the LangChain framework \citep{langchain} for this purpose. This recursive approach was adopted to ensure optimal semantic coherence by prioritizing natural delimiters (e.g., paragraphs and sentences). \begin{equation}
    X \xrightarrow{\text{Split}} C = (C_1, C_2, \dots, C_K).
\end{equation}

Given a query $Q$, we append a fixed number of \texttt{<|CPS|>} tokens to each chunk $C_j$ and process them in parallel with the knowledge model $\psi_{\text{kno}}$, using the last-layer hidden states of these compression tokens as the latent representation $T_j$.
\begin{equation}
    \psi_{\text{kno}}:(C_j, Q) \;\rightarrow\; T_j \in \mathbb{R}^{\xi_j \times d}.
\end{equation}

Finally, the outputs from all chunks are concatenated in the original order to yield the global sequence of implicit fact tokens, denoted $T$.
\begin{equation}
T = \mathrm{Concat}(T_1, \dots, T_K)
  = [t_1, \dots, t_{\xi}] \in \mathbb{R}^{\xi \times d},
\end{equation}
\noindent
where $\xi = \sum_{j=1}^{K} \xi_j \ll N$.

Implementation Note: During concatenation, a double newline separator (\texttt{\textbackslash n\textbackslash n}) is inserted between adjacent sub-sequences $T_j$ and $T_{j+1}$ to mark chunk boundaries.
We define \(T_i \in \mathbb{R}^d\) as implicit fact tokens, and d denotes the hidden state dimensionality. These tokens encapsulate essential information from the input and serve as continuous latent units for the reasoning model.

In the next step, a three-layer MLP projector $\pi$ maps the implicit fact tokens into implicit fact embeddings $E = (e_1, e_2, \ldots, e_{\xi})$, thereby aligning them with the embedding space of the reasoning model. These embeddings, together with the query embeddings $E(Q)$, are subsequently fed into the reasoning model for downstream inference. 
\begin{equation}
    \theta_{\text{rea}} : \mathrm{Concat}(E, E(Q)) \;\;\rightarrow\;\; \text{Response}
\end{equation}                                       
The Response denotes the thoughts and the final answer generated by the reasoning model.

This approach not only substantially reduces GPU memory consumption but also frees the reasoning model from processing lengthy documents filled with irrelevant information.

\subsection{Task Definition}
\begin{figure*}
    \centering
    \includegraphics[width=1\linewidth]{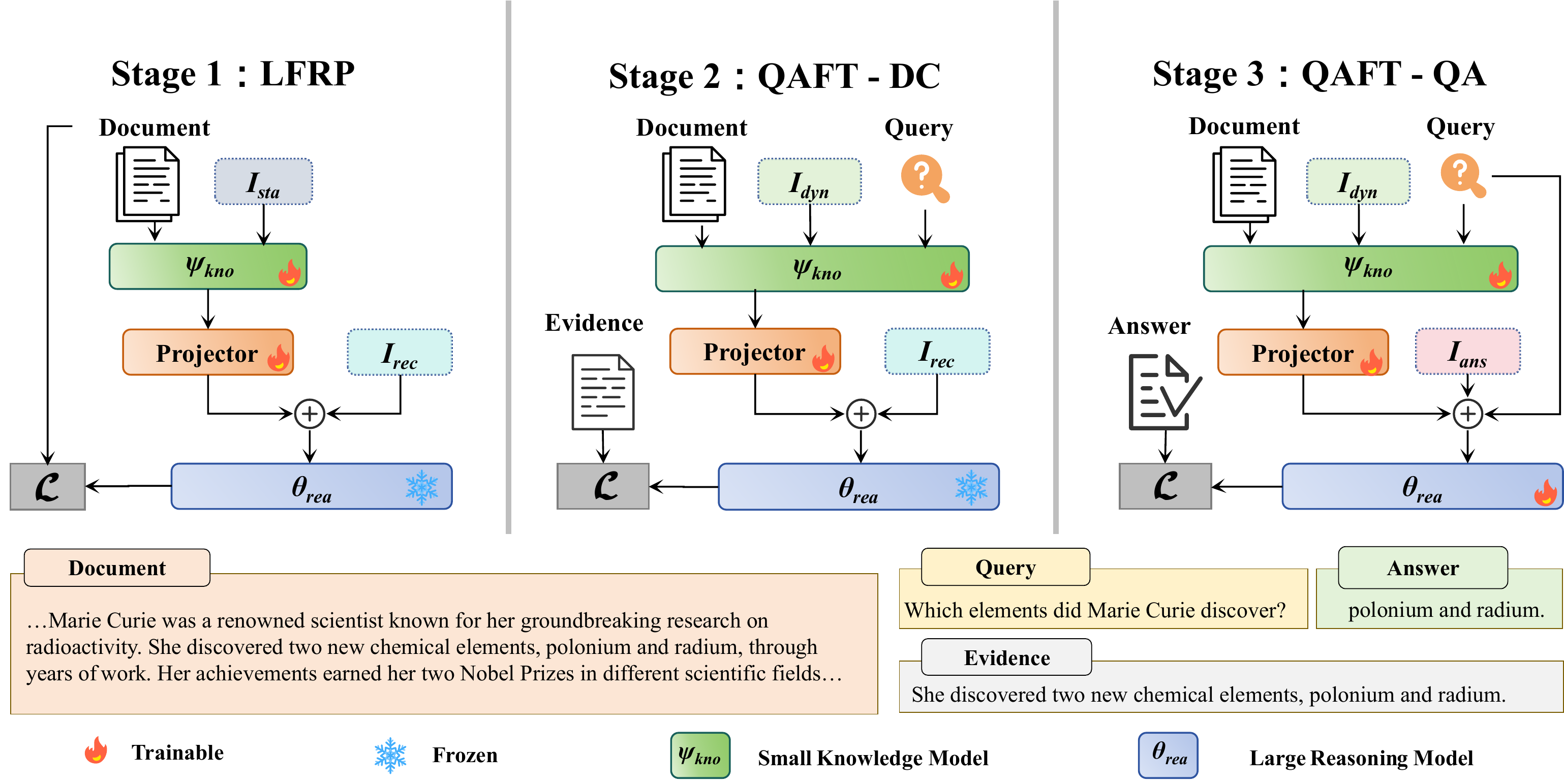}
    \caption{Three different trainging tasks for DRIFT. The instructions in the figure include the dynamic compression instruction, reconstruct instruction, answer instruction, and static compression instructio.}
    \label{fig:train}
\end{figure*}

To more effectively achieve the decoupling of knowledge and reasoning, we decompose the training of DRIFT into three distinct stages, each optimized for a different objective, as shown in Figure \ref{fig:train}.
\subsubsection{Latent Fact Reconstruction Pretraining (LFRP) }
We redefine the pretraining objective such that the reasoning model is used only as a frozen decoder to provide a reconstruction signal, while the knowledge model is optimized to generate latent factual representations that best support document reconstruction.

For the knowledge model, this compression is static because it is query-independent. We set the static compression ratio $c_{\text{sta}}$ to 8. After applying the bucketed compression strategy, we obtain $\xi_{\text{sta}}$, denoting the number of implicit fact tokens.

Given a document $x$ consisting of $n$ tokens, $X = (x_1, x_2, \ldots, x_n)$, the knowledge model produces latent fact tokens conditioned on a static compression instruction $I_\text{sta}$:
\begin{equation}
T_{\text{sta}} = [t_1, t_2, \ldots, t_{\xi_{\text{sta}}}] = \psi_{\text{kno}}(I_\text{sta}, x_1, x_2,\ldots, x_N).
\end{equation}

These implicit fact tokens are projected into fact embeddings via a projector module $\pi$:
\begin{equation}
E_{\text{sta}} = \pi(T_{\text{sta}}) = (e_1, e_2, \ldots, e_{\xi_{\text{sta}}}).
\end{equation}

The reasoning model, parameterized by $\theta_r$, remains frozen during this pretraining stage. It receives a reconstruction instruction $I_\text{rec}$ and predicts each token conditioned on the fact embeddings and previously generated tokens:
\begin{equation}\small
\mathcal{L}(\psi_{\text{kno}}, \pi)
= -\sum_{x_t \in X} \log P_{\theta_{\text{rea}}}(x_t \mid I_\text{rec}, E, x_{<t}),
\end{equation}

Thus, although the loss is computed using the frozen reasoning model, gradients are only backpropagated through $E_{\text{sta}}$ into the projector $\pi$ and the knowledge model $\psi_\text{kno}$. This teaches the knowledge model to produce latent factual representations that are maximally useful for reconstructing the original document.

\subsubsection{Query-Aware Fine-Tuning (QAFT) with Single-Context}
Through the pretraining objective described above, the knowledge model acquires the ability to encode factual knowledge into a latent space, while the reasoning model learns to understand the implicit fact embeddings. To further adapt the framework for downstream tasks, we introduce an additional training objective that incorporates query inputs. This objective is designed to encourage the knowledge model to implicitly extract and compress query-relevant knowledge into the latent space, and to enable the reasoning model to perform question answering by leveraging the information encoded in the implicit fact embeddings. We train the models on QA datasets, including reading comprehension, fact verification and open-domain question answering.  

To better train toward this objective, we perform two fine-tuning tasks in sequence: first a dynamic compression task, followed by a question-answering task.

\paragraph{Dynamic Compression Task}
In the pretraining task, the knowledge model learns to perform static compression of the context. Now, we aim to train the model to acquire query-aware dynamic compression capability.

In training dataset, each question-answer pair is annotated with supporting evidence. We design a dynamic compression task using the context and question-evidence pairs: a knowledge model extracts query-specific information from the context into a latent space (implicit fact tokens), while a reasoning model reconstructs the evidence from these tokens. By leveraging the sparsity of query-relevant information within the context, dynamic compression can safely achieve a significantly higher compression ratio than static approaches. The default dynamic compression ratio is set to 32.

We use the instruction $I_\text{dyn}$ to ask the knowledge model to  extract question-relevant knowledge from the document and encodes it into a set of implicit fact tokens in the latent space. 
Conditioned on the input query Q, the knowledge model $\psi_{\text{kno}}$ generates a sequence of question-aware implicit fact tokens $T_{\text{dyn}}$. 

The reconstruction performed by the reasoning model under instruction $I_\text{rec}$ remains analogous to the pre-training stage. However, we diverge by restricting the reconstruction target to the evidence instead of the complete original text. This mechanism employs the evidence as the supervisory signal, explicitly training the knowledge model to develop dynamic compression capabilities.

Given a document $X = (x_1, x_2, \ldots, x_n)$ and a question $Q = (q_1, q_2, \ldots, q_m)$, 
the knowledge model $\psi_k$ performs dynamic compression on $X$ conditioned on $Q$.:
\begin{equation}
    T_{\text{dyn}} = (t_1, t_2, \ldots, t_{\xi_{\text{dyn}}}) = \psi_{\text{kno}}(I_{\text{dyn}}, X, Q)
\end{equation}

These latent tokens are then projected into the final fact embeddings $E_{\text{dyn}} = (e_1, e_2, \ldots, e_{\xi_{\text{dyn}}})$ via the projector module $\pi$. The reasnoning model still keeps fixed and is asked to reconstruct the evidence $X_{\text{evi}}$:
\begin{equation}
\small
\mathcal{L}(\psi_{\text{kno}}, \pi)
= -\sum_{x_k \in X_{\text{evi}}} \log P_{\theta_{\text{rea}}}(x_k \mid I_{\text{rec}}, E_{\text{dyn}}, x_{<k})
\end{equation}

\paragraph{Question-answering Task}
This task is the only one in which the reasoning model is not frozen during training, enabling it to better exploit the compressed context for downstream tasks. We perform fine-tuning by updating the models solely based on the target answers. 

The generation of $E_{\text{dyn}}$ leverages the same mechanism as the Dynamic Compression Task, implemented by the Knowledge model and projector. Then we instruct the reasoning model with $I_\text{ans}$ to generate the answer of the question based on the fact embeddings. We denote the answer sequence as $A = (a_1, a_2, \ldots, a_l)$.

We define the training objective as the standard language modeling loss. Our formulation is largely analogous to instruction tuning \citep{instruction_tuning}, with the key distinction that the context presented to the reasoning model is transformed from the explicit text tokens to the implicit fact embeddings produced by the knowledge model. The question embedding is represented by $E(Q)$. We minimize the following loss to optimize the model parameters:
\begin{equation}\small
\mathcal{L}(\theta_{\text{rea}}, \psi_{\text{kno}})
= -\sum_{a_j \in A} \log P_{\theta_{\text{rea}}}
\bigl(a_j \mid I_{\text{ans}}, E_{\text{dyn}}, E(Q), a_{<j}\bigr)
\end{equation}

\subsection{Multi-Context Inference without Fine-Tuning}
We find that DRIFT, fine-tuned solely in the single-context setting via QAFT, generalizes effectively to multi-context inference without requiring any additional multi-context training.
Specifically, to process extensive contexts, we employ an overlapping chunking strategy utilizing \texttt{RecursiveCharacterTextSplitter}. We set the chunk size to 8,192 tokens, aligning with the maximum document length used during training. These chunks are then dynamically compressed in parallel by the knowledge model.

\subsection{Training Data}
We use the English Wikipedia snapshot dated November 1, 2023, treating each entry as a document and its text field as the raw training content. Since longer input documents make model training and convergence more challenging, we adopt a token-level curriculum learning strategy across all three training tasks, with phases defined by input document length. Details of the construction procedures for LFRP and QAFT, as well as the curriculum learning partitions, are provided in Appendix \ref{app:data_construct}.

\section{Experiments}
\subsection{Implement Details}
For the DRIFT setup, we fix \texttt{Qwen2.5-Instruct-3B} as the knowledge model and evaluate reasoning models from different families, including Mistral\citep{mistral} and Qwen2.5\citep{qwen25}, with \texttt{Mistral-7B-Instruct-v0.2} as default. We further explore additional configurations, such as smaller knowledge models (1.5B) and larger reasoning models (14B). All models are trained using parameter-efficient fine-tuning with LoRA \citep{lora}. The results for broader model combinations, along with detailed training hyperparameters, are provided in the appendix.

\smallskip
\noindent
\textbf{RQ1: To what extent do the latent representations capture and preserve the essential information of the original context?}
\smallskip

To assess whether the knowledge model can compress long contexts with minimal information loss, we evaluate the reconstruction fidelity of the learned latent representations before downstream reasoning. Specifically, after training on the LFRP task, we conduct a compression–reconstruction experiment on a test subset with input lengths ranging from 512 to 1024 tokens, where the reasoning model reconstructs the original text conditioned solely on the compressed representations produced by the knowledge model. Reconstruction quality is measured using BLEU and ROUGE (ROUGE-1/2/L) scores.

\begin{table}[htbp]
\centering
\small
\setlength{\tabcolsep}{4pt}
\resizebox{\linewidth}{!}{ 
\begin{tabular}{ll|cccc}
\toprule
\multicolumn{2}{l|}{\textbf{Models}} & \textbf{BLEU} & \textbf{R-1} & \textbf{R-2} & \textbf{R-L} \\
\midrule
\multirow{2}{*}{\textbf{DRIFT after LFRP}} & Mistral-7B & 90.01 & 94.95 & 93.93 & 94.75 \\
 & Qwen2.5-7B & 83.43 & 86.60 & 84.13 & 88.80 \\
\midrule
\multirow{2}{*}{\textbf{DRIFT before LFRP}} & Mistral-7B & 0.01 & 1.63 & 0.07 & 1.46 \\
 & Qwen2.5-7B & 0.00 & 6.32 & 0.65 & 4.52 \\
\bottomrule
\multicolumn{6}{l}{\footnotesize \textit{Note:} R-1/2/L stands for ROUGE-1/2/L scores. Scores are scaled by 100.} 
\end{tabular}
}
\caption{Performance evaluation of the reconstruction task across different model configurations.}
\label{tab:LFRP_eval}
\end{table}

As illustrated in Table \ref{tab:LFRP_eval}, the compression–reconstruction task achieves strong performance across all model combinations following the LFRP training phase. This indicates that after training, the knowledge model effectively compresses the input content and the reasoning model accurately interprets the compressed representations.

\begin{table*}[!htbp]
\centering
\renewcommand{\arraystretch}{1.3} 
\setlength{\tabcolsep}{2.5pt} 

\resizebox{\textwidth}{!}{%
\begin{tabular}{@{} l c @{\hspace{5pt}} ccc g @{\hspace{5pt}} ccc g @{\hspace{5pt}} cc g @{\hspace{5pt}} cccc g @{\hspace{5pt}} ccc g @{}}
\toprule

\multirow{3}{*}{\textbf{Model}} 
& \multirow{3}{*}{\textbf{\shortstack{Comp. \\ Ratio}}} 
& \multicolumn{4}{c}{\textbf{BAMBOO (16k)}} 
& \multicolumn{4}{c}{\textbf{L-Eval (QA Subset)}} 
& \multicolumn{3}{c}{\textbf{L-Eval (Sum Subset)}} 
& \multicolumn{5}{c}{\textbf{LoCoMo}} 
& \multicolumn{4}{c}{\textbf{LongBench-v2}} \\

\cmidrule(lr){3-6} \cmidrule(lr){7-10} \cmidrule(lr){11-13} \cmidrule(lr){14-18} \cmidrule(l){19-22}

 & & \multirow{2}{*}{AltQA} & \multirow{2}{*}{Meet} & \multirow{2}{*}{Paper} & \raisebox{-1.5ex}{\textbf{Avg}}
 & \multirow{2}{*}{NQ} & \multirow{2}{*}{NarQA} & \multirow{2}{*}{Crse} & \textbf{Avg} 
 & \multirow{2}{*}{QMS} & \multirow{2}{*}{SPC} & \textbf{Avg} 
 & \multirow{2}{*}{T1} & \multirow{2}{*}{T2} & \multirow{2}{*}{T3} & \multirow{2}{*}{T4} & \textbf{Avg} 
 & \multirow{2}{*}{Short} & \multirow{2}{*}{Medium} & \multirow{2}{*}{Long} & \textbf{Avg} \\

 & & & & & & & & & & & & & & & & & & & & & \\

\midrule

\multicolumn{22}{@{}l}{\textit{\textbf{Baseline Methods based on Mistral-7B-v0.2}}} \\
\midrule

LLMLingua-2 & 3$\times$ & 40.00 & 75.00 & 76.77 & \cellcolor{gray!25}57.89 & 77.98 & 33.18 & 64.53 & \cellcolor{gray!25}53.94 & 19.37 & 18.86 & \cellcolor{gray!25}19.14 & 46.81 & 27.41 & 48.96 & 76.93 & \cellcolor{gray!25}59.35 & 26.11 & 13.95 & 25.00 & \cellcolor{gray!25}20.68 \\
NaiveRAG & -- & 34.50 & 79.00 & 75.76 & \cellcolor{gray!25}55.89 & 72.48 & 21.50 & 63.37 & \cellcolor{gray!25}47.27 & 19.61 & 18.11 & \cellcolor{gray!25}18.94 & 45.04 & 24.61 & 41.67 & 73.48 & \cellcolor{gray!25}56.10 & 17.78 & 26.05 & 25.00 & \cellcolor{gray!25}22.86 \\
xRAG & 128$\times$ & 25.00 & 68.00 & 57.58 & \cellcolor{gray!25}43.86 & 56.88 & 12.62 & 50.58 & \cellcolor{gray!25}35.56 & 18.06 & 12.34 & \cellcolor{gray!25}15.49 & 20.57 & 6.23 & 34.38 & 32.10 & \cellcolor{gray!25}24.74 & 31.11 & 25.58 & 25.00 & \cellcolor{gray!25}27.43 \\
ICAE & 4$\times$ & 21.50 & 19.00 & 19.19 & \cellcolor{gray!25}20.30   & 36.70 & 5.14 & 15.70 & \cellcolor{gray!25}15.76 & 17.62   & 13.27 & \cellcolor{gray!25}15.67 & 17.38 & 3.43 & 23.96 & 15.10 & \cellcolor{gray!25}13.64 & 13.89 & 14.88 & 10.56 & \cellcolor{gray!25}13.12 \\
\rowcolor{gray!5} 
COCOM & 4$\times$ & 30.50 & 60.00 & 47.47 & \cellcolor{gray!25}42.11 & 61.47 & 9.81 & 53.49 & \cellcolor{gray!25}36.36 & 9.15 & 7.79 & \cellcolor{gray!25}8.54 & 9.93 & 1.87 & 27.08 & 10.34 & \cellcolor{gray!25}9.55 & 27.22 & 23.26 & 34.26 & \cellcolor{gray!25}27.04 \\
\rowcolor{gray!5}
 & 16$\times$ & 25.50 & 49.00 & 41.41 & \cellcolor{gray!25}35.34 & 53.21 & 8.88 & 55.81 & \cellcolor{gray!25}34.95 & 7.78 & 5.37 & \cellcolor{gray!25}6.70 & 11.35 & 1.87 & 25.00 & 10.11 & \cellcolor{gray!25}9.55 & 27.78 & 23.26 & 34.26 & \cellcolor{gray!25}27.24 \\
\rowcolor{gray!5}
 & 128$\times$ & 15.00 & 40.00 & 36.36 & \cellcolor{gray!25}26.57 & 40.37 & 5.61 & 39.53 & \cellcolor{gray!25}25.05 & 8.71 & 4.79 & \cellcolor{gray!25}6.95 & 12.06 & 1.56 & 30.21 & 8.32 & \cellcolor{gray!25}8.96 & 29.44 & 21.40 & 34.26 & \cellcolor{gray!25}27.04 \\
\midrule

\multicolumn{22}{@{}l}{\textit{\textbf{DRIFT based on Mistral-7B-v0.2}}} \\
\midrule
\textbf{DRIFT (Ours)} & \textbf{32$\times$} & \textbf{41.50} & \textbf{80.00} & \textbf{77.78} & \cellcolor{gray!25}\textbf{60.15} & \textbf{69.72} & \textbf{27.10} & \textbf{61.05} & \cellcolor{gray!25}48.28 & \textbf{21.59} & \textbf{19.75} & \cellcolor{gray!25}20.76 & \textbf{36.88} & \textbf{23.99} & \textbf{33.33} & \textbf{80.38} & \cellcolor{gray!25}\textbf{57.73} & \textbf{32.22} & \textbf{28.84} & \textbf{25.00} & \cellcolor{gray!25}\textbf{29.22} \\
\textbf{} & \textbf{64$\times$} & \textbf{36.00} & \textbf{82.00} & \textbf{82.83} & \cellcolor{gray!25}\textbf{59.15} & \textbf{70.64} & \textbf{24.77} & \textbf{58.72} & \cellcolor{gray!25}46.67 & \textbf{21.95} & \textbf{19.49} & \cellcolor{gray!25}20.85 & \textbf{34.75} & \textbf{16.82} & \textbf{42.71} & \textbf{74.67} & \cellcolor{gray!25}\textbf{53.31} & \textbf{26.67} & \textbf{29.77} & \textbf{19.44} & \cellcolor{gray!25}\textbf{26.44} \\
\textbf{} & \textbf{128$\times$} & \textbf{36.00} & \textbf{77.00} & \textbf{78.79} & \cellcolor{gray!25}\textbf{56.89} & \textbf{71.56} & \textbf{17.29} & \textbf{60.47} & \cellcolor{gray!25}44.24 & \textbf{20.69} & \textbf{19.14} & \cellcolor{gray!25}19.99 & \textbf{30.14} & \textbf{17.13} & \textbf{36.46} & \textbf{72.06} & \cellcolor{gray!25}\textbf{50.71} & \textbf{32.22} & \textbf{24.65} & \textbf{19.44} & \cellcolor{gray!25}\textbf{26.24} \\
\midrule
\multicolumn{22}{@{}l}{\textit{\textbf{DRIFT based on Qwen2.5-Instruct-7B}}} \\
\midrule
\textbf{DRIFT (Ours)} & \textbf{32$\times$} & \textbf{37.00} & \textbf{81.00} & \textbf{84.85} & \cellcolor{gray!25}\textbf{59.90} & \textbf{80.73} & \textbf{27.10} & \textbf{76.16} & \cellcolor{gray!25}55.96 & \textbf{22.66} & \textbf{19.01} & \cellcolor{gray!25}21.02 & \textbf{42.55} & \textbf{27.73} & \textbf{40.62} & \textbf{85.49} & \cellcolor{gray!25}\textbf{62.79} & \textbf{36.67} & \textbf{28.37} & \textbf{33.33} & \cellcolor{gray!25}\textbf{32.41} \\
\midrule
\multicolumn{22}{@{}l}{\textit{\textbf{Vanilla LLM}}} \\
\midrule
Mistral-7B-v0.2 & 1$\times$ & 40.00 & 75.00 & 76.77 & \cellcolor{gray!25}57.89 & 79.82 & 31.78 & 65.12 & \cellcolor{gray!25}53.94 & 19.32 & 18.90 & \cellcolor{gray!25}19.13 & 46.81 & 27.73 & 48.96 & 76.81 & \cellcolor{gray!25}59.35 & 25.00 & 16.28 & 23.15 & \cellcolor{gray!25}20.87 \\
Qwen2.5-Instruct-7B & 1$\times$ & 36.00 & 78.00 & 82.83 & \cellcolor{gray!25}58.15 & 80.73 & 33.18 & 75.58 & \cellcolor{gray!25}58.38 & 20.74 & 18.50 & \cellcolor{gray!25}19.74 & 48.23 & 38.01 & 46.88 & 85.14 & \cellcolor{gray!25}66.17 & 41.11 & 26.05 & 24.07 & \cellcolor{gray!25}31.01 \\
\bottomrule
\end{tabular}%
}
\caption{Main results of DRIFT and other baseline methods on long context datasets.}
\label{tab:main_results}
\end{table*}
\smallskip
\noindent
\textbf{RQ 2: How does DRIFT compare to representative baselines in terms of overall effectiveness in long-context reasoning scenarios?}
\smallskip

\paragraph{Benchmarks} To evaluate DRIFT across diverse long-context scenarios, we employ several representative benchmarks. \textbf{BAMBOO} \citep{bamboo} serves as a comprehensive suite for testing extended context capabilities through tasks like question answering and code completion. \textbf{L-Eval} \citep{leval} provides a balanced evaluation of single-hop and multi-hop reasoning across varying document lengths up to 256K tokens while minimizing knowledge leakage. \textbf{LongBench-v2} \citep{longbenchv2} consists of challenging multiple-choice questions requiring multi-document understanding and structured data reasoning across contexts reaching millions of words. Finally, \textbf{LoCoMo} \citep{locomo} focuses on long-term conversational memory, probing the model's ability to perform temporal reasoning and event summarization over multi-session dialogue histories.

\paragraph{Metrics}
There are two primary metrics. For datasets involving multiple-choice, closed-ended, and measurable open-ended tasks, we utilize \texttt{Qwen-2.5-72B-Instruct} as an \textbf{LLM-Judge} to compute accuracy. For summarization-oriented tasks, we report \textbf{ROUGE-L} scores to measure the similarity between generated responses and ground-truth references based on the Longest Common Subsequence.

\paragraph{Baselines}
We compare \textbf{DRIFT} against a diverse set of baseline methods categorized by their compression and retrieval paradigms. For \textbf{hard compression}, we select \textbf{LLMLingua-2} to represent lexical-level token pruning. For \textbf{soft compression}, we evaluate several latent-space models including \textbf{ICAE}, \textbf{COCOM}, and \textbf{xRAG}. Additionally, we implement a \textbf{NaiveRAG} baseline utilizing the \textbf{BGE-M3} embedding model for document retrieval. The \textbf{Mistral-7B-v0.2} model serves as our primary vanilla backbone to provide a performance lower bound without external enhancements. 

\paragraph{Results and Analysis}
Table \ref{tab:main_results} demonstrates that DRIFT consistently outperforms existing compression-based baselines across diverse long-context benchmarks, particularly under high compression ratios.
With the same Mistral backbone, DRIFT sets a new state of the art over existing compression-based methods. Notably, on task-oriented summarization benchmarks (\textbf{QMSUM} and \textbf{SPACE}) and the \textbf{LoCoMo} conversational memory benchmark—where prior compression approaches largely fail—DRIFT remains effective, highlighting its superior ability to preserve and exploit long-range contextual information.

\smallskip
\noindent
\textbf{RQ3: How does each training objective contribute to the overall performance of DRIFT?}

DRIFT incorporates three distinct training objectives: LFRP, QAFT-DC, and QAFT-QA. To justify the necessity of this multi-stage design and quantify the contribution of each component, we conduct a comprehensive ablation study across three long-context benchmarks.

\begin{table}[htbp]
\centering
\small
\setlength{\tabcolsep}{4pt}
\begin{tabularx}{\linewidth}{l c c c}
\toprule
\textbf{Method} 
& \textbf{Bamboo} 
& \textbf{LongBenchv2} 
& \textbf{LoCoMo} \\
\midrule
DRIFT (32$\times$)
& \textbf{60.15} & \textbf{29.22} & \textbf{57.73} \\
w/o LFRP 
& 57.64$^{\downarrow 2.51}$ 
& 26.84$^{\downarrow 2.38}$ 
& 52.68$^{\downarrow 5.05}$ \\
w/o QAFT-DC 
& 56.64$^{\downarrow 3.51}$ 
& 25.84$^{\downarrow 3.38}$ 
& 57.36$^{\downarrow 0.37}$ \\
w/o QAFT-QA 
& 45.14$^{\downarrow 15.01}$ 
& 18.05$^{\downarrow 11.17}$ 
& 36.89$^{\downarrow 20.84}$ \\
\bottomrule
\bottomrule
\multicolumn{4}{p{\linewidth}}{\footnotesize \textit{Note:} QAFT-DC denotes the \textit{QAFT Dynamic Compression} objective, and QAFT-QA denotes the \textit{QAFT Question Answering} objective.}
\end{tabularx}
\caption{
DRIFT ablation results (avg. accuracy).
}\label{tab:ablation}
\end{table}

As shown in Table \ref{tab:ablation}, each training objective in DRIFT serves a unique purpose: QAFT-QA provides the fundamental reasoning backbone, while LFRP and QAFT-DC further optimize the latent space efficiency and robustness.

\smallskip
\noindent
\textbf{RQ4: Does DRIFT maintain competitive inference efficiency compared to classical baselines?}

Efficiency is a critical bottleneck for long-context reasoning. We conduct an end-to-end Time-to-First-Token (TTFT) analysis to evaluate whether DRIFT maintains its performance advantages without incurring prohibitive computational costs as the input scale grows. The end-to-end Token-to-First-Token (TTFT) measures the latency from providing both the input documents and the query to the system until the first output token is generated. This metric directly captures the practical responsiveness of long-context reasoning systems under realistic inference settings.

\begin{figure}[htbp]
    \centering
    \includegraphics[width=0.95\linewidth]{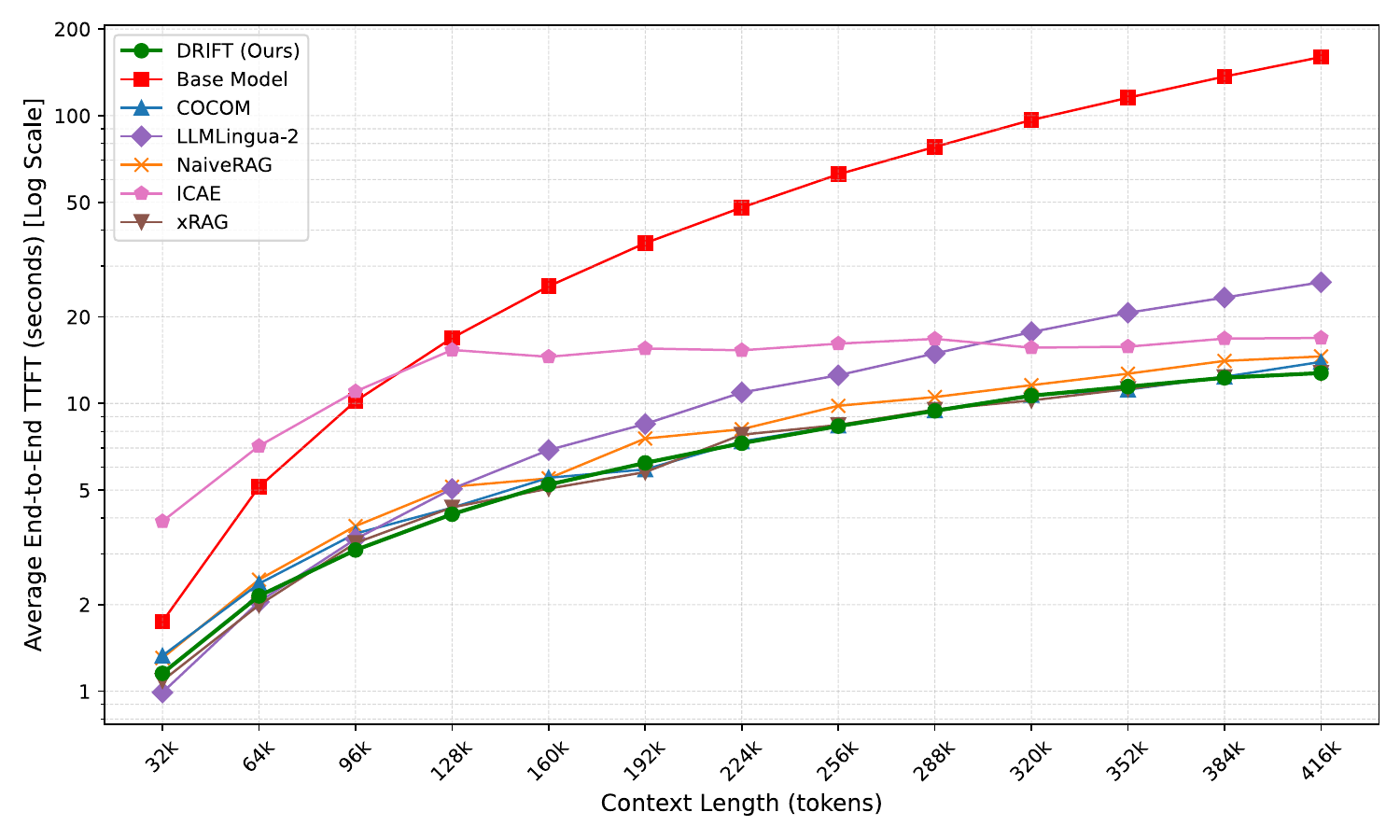}
    \caption{End-to-end TTFT as a function of input length for different baselines.}
    \label{fig:ttft}
\end{figure}

Figure~\ref{fig:ttft} shows that DRIFT maintains competitive efficiency across all baselines. Notably, the performance gap between DRIFT and the \textit{Full-Context} baseline widens significantly as the input length increases, demonstrating DRIFT's superior scalability for ultra-long sequences.

\section{Conclusion}
In this paper, we introduced \textbf{DRIFT}, a dual-model architecture designed to decouple factual knowledge acquisition from general reasoning in LLMs. Input documents are segmented into chunks, compressed into high-density fact embeddings by a lightweight knowledge model, and interpreted by a larger reasoning model. Experimental results and ablation studies demonstrate the effectiveness of DRIFT and the necessity of each training task. The method also generalizes well across different compression ratios and backbone models, highlighting its robustness and practical applicability.
\section{Limitations}
Despite its effectiveness, our work has certain limitations that suggest directions for future research. First, due to computational resource constraints, our experiments were primarily validated on models with up to 14B parameters; the performance and scaling laws of DRIFT on larger-scale models remain to be further explored. Second, the use of latent compression introduces challenges regarding interpretability, as the implicit fact tokens are not as human-readable as raw text snippets in traditional RAG. Finally, our current framework is primarily optimized through Supervised Fine-Tuning (SFT). We anticipate that integrating Reinforcement Learning (RL) could further enhance the model's decision-making in knowledge selection and lead to new breakthroughs in performance.

\bibliography{custom}
\clearpage
\newpage
\appendix
\section{Data Generation Details} \label{app:data_construct}
We compute token lengths for Wikipedia documents and sample text segments for each length bucket. Specifically, for the LFRP task, we sample 200,000 segments per bucket, ensuring full coverage including the 4k8k range. For the QA-FT task, we sample 100,000 segments per bucket, with the exception of the 4k–8k range where data remains insufficient. Each stage is then split into training, validation, and test sets with an 8:1:1 ratio.
\subsection{LFRP Data}
For the unlabeled data used in pre-training, we deliberately refrain from any data cleaning. Since documents or knowledge bases in real-world scenarios are often noisy and heterogeneous in format, we retain the raw text to improve the robustness of the pre-trained model. For each stage, we use 160,000 samples to train the model.
\subsection{QAFT Data}
\begin{figure*}
    \centering
    \includegraphics[width=1\linewidth]{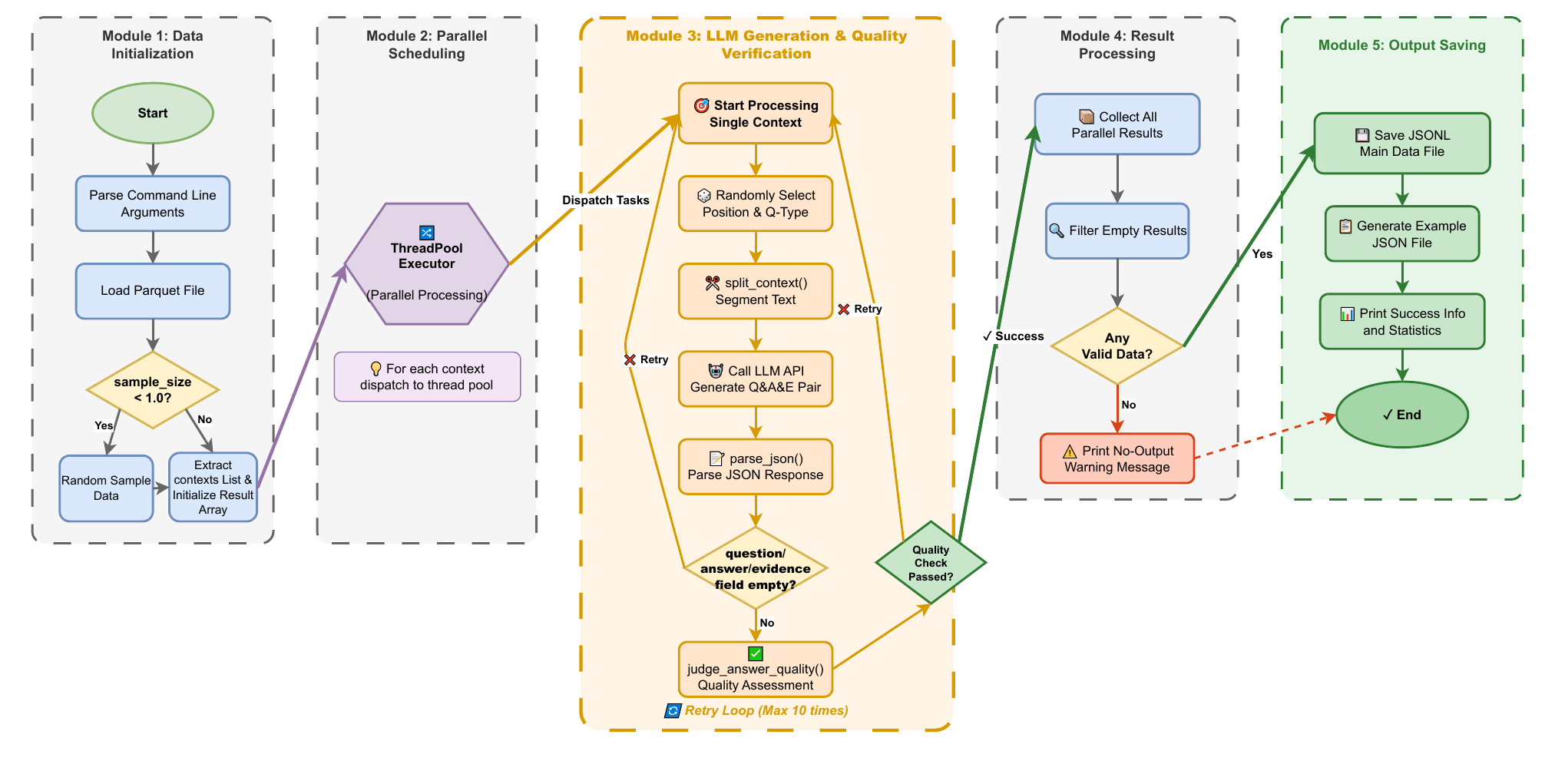}
    \caption{An Automated Pipeline for Contextual Question-Answering Data Synthesis}
    \label{fig:data_gen}
\end{figure*}

Based on raw Wikipedia documents corresponding to each length bucket, we employed Qwen2.5-72B-Instruct to generate QA pairs, 
which were subsequently used for fine-tuning in each training stage. In general, each document corresponds to one QA pair; however, for the 4k–8k stage, the available raw documents were insufficient, so some documents were reused. To avoid positional bias in the generated QA data, we segment each document and randomly sample one slice as the input for QA generation. To avoid positional bias in the generated QA data, we segment each document and randomly sample one slice as the input for QA generation. The workflow of data generation is shown in Figure \ref{fig:data_gen}.

\paragraph{Data Sampling Strategy}
To ensure that relevant information is uniformly distributed across the generated dataset, we adopt the following sampling strategy. Each Wikipedia document is first divided into multiple slices of equal length. For every document, we then randomly select one slice as the context for QA generation. This prevents the model from always encountering answers concentrated in specific regions (e.g., the beginning of documents) and ensures that relevant content appears at random positions. As a result, the generated QA pairs cover diverse locations within documents, leading to a more balanced and robust training signal. In addition to generating the question–answer pairs, the model is also required to provide the supporting evidence from the original text corresponding to each answer.

\paragraph{Question Type Diversification}
To better train the reasoning model to utilize information compressed by the knowledge model, we randomly sample one of three formats—multiple-choice, true/false, or short-answer questions—for QA generation. This exposes the model to diverse reasoning scenarios, thereby strengthening its ability to integrate knowledge-derived information for effective problem solving, and enhancing its generalization across different task settings.

\paragraph{Data Filtering}
To ensure the quality of the automatically generated QA pairs, we employ Qwen2.5-72B-Instruct as the judger model to filter out low-quality instances. Specifically, the filtering process is guided by five criteria:
\begin{enumerate*}[label=(\roman*)]
    \item \textbf{Relevance}: the question must be grounded in the given context;
    \item \textbf{Correctness}: the provided answer should be factually consistent with the context;
    \item \textbf{Clarity}: the question and answer must be well-formed and unambiguous;
    \item \textbf{Fidelity}: the evidence must accurately reflect content from the original document without introducing external information;
    \item \textbf{Sufficiency}: the evidence must provide enough information to answer the question.
\end{enumerate*}
Only QA pairs that satisfy all conditions are retained for subsequent training.

\subsection{Training Strategy: Token-Level Curriculum Learning}
During model training, we observed that longer input documents lead to increased training difficulty. In particular, the convergence of both the knowledge and reasoning models becomes more challenging as the number of tokens grows. Therefore, we apply a \textbf{token-level curriculum learning strategy} across all three training tasks, where training stages are defined according to the token length of the input documents. Table \ref{tab:curriculum} summarizes the token-level stage configuration for each of the three training tasks.

\begin{table}[htbp]
\centering
\resizebox{\linewidth}{!}{
\begin{tabular}{lcccc}
\toprule
\textbf{Task} & \textbf{Stage 1} & \textbf{Stage 2} & \textbf{Stage 3} & \textbf{Stage 4} \\
\midrule
\textit{\textbf{LFRP}} & 64-128 & 128-256 & 256-512 & 512-1k \\
\midrule
\multicolumn{5}{l}{\textit{\textbf{QAFT Tasks}}} \\ 
\hspace{1em} Dynamic Comp. & 1k-2k & 2k-4k & 4k-8k & -- \\
\hspace{1em} Question Ans. & 1k-2k & 2k-4k & 4k-8k & -- \\
\bottomrule
\end{tabular}
}
\caption{Curriculum learning stages. Note that QAFT tasks conclude at Stage 3.}
\label{tab:curriculum}
\end{table}

\subsection{Construction Prompt}
Two distinct prompts serve data construction: one for generation and one for fine-grained evaluation.
\begin{promptbox}
    \textbf{Prompt 1} ``
    Please generate a question that can be answered based on the provided context. The question should be highly relevant to the context, and the answer must be directly inferable from the given information. Avoid asking questions that cannot be answered using the context. The question should be of the type: \{question\_type\}.

    Your response should consist of three parts:
    
1. Question – the generated question. (a string)

2. Answer – the answer, including how it is reasoned out from the relevant information in the context.  (a string)

3. Evidence – the specific part(s) of the original text that support the answer.  (a string)

Attention: Evidence must be quoted directly from the original text and must include all the information needed to answer the question. If some parts of the evidence involve unclear references (e.g., ambiguous subjects), include the related sentences that clarify them, so that the evidence alone is sufficient for answering the question. Ensure that every sentence remains complete, without the use of ellipses.

Your output format should be:

json

\{\{

    "question": "<the generated question (include options if the question type is multiple choice)>",
    
    "answer": "<the corresponding answer, including how it is inferred from the relevant information in the context>",
    
    "evidence": "<the specific part(s) taken directly from the original text that support the answer>"
    
\}\}

Context: \{context\}

Your output: '',
\end{promptbox}
\begin{promptbox}
    \textbf{Prompt 2} ``You are a judge evaluating the quality of question-answer pairs. Your task is to determine whether the given answer can be reasonably inferred from the provided evidence.

Please evaluate based on the following criteria:

1. Can the answer be directly supported by the evidence?

2. Is the evidence sufficient to answer the question?

3. Is the answer logically consistent with the evidence?

4. Are there any contradictions between the answer and evidence?

Question: \{question\}

Evidence: \{evidence\}

Answer: \{answer\}

Please respond with only "true" if the answer can be reasonably inferred from the evidence, or "false" if it cannot.

Your judgment:'',
\end{promptbox}

\section{Dataset Statistics}
\label{sec:appendix_data_stats}
Dataset Statistics and Configuration: Table \ref{tab:token_statistics} summarizes the statistics of our cleaned raw dataset. The dataset comprises approximately 2.13 million samples with a total of 1.61 billion tokens, covering a wide range of sequence lengths from 64 to 4,096. This diverse distribution ensures the model's robustness across various context windows. Our final datasets were constructed by extracting and processing samples from this original dataset. For the training pipeline, we strategically allocated the data across different phases: the LFRP stage utilized 640,000 samples to establish solid feature representations, while the QAFT stage employed 240,000 specifically constructed samples to refine the model's task-specific performance.

\begin{table}[t]
\centering
\caption{Statistics of the dataset across different token length ranges.}
\label{tab:token_statistics}
\fontsize{10pt}{11pt}\selectfont
\begin{tabular}{llrr}
\toprule
\textbf{Token Range} & \textbf{Split} & \textbf{Total Tokens} & \textbf{Samples} \\ \midrule
\multirow{4}{*}{64 -- 128}   & Train & 30,511,507 & 320,000 \\
                             & Val   & 3,813,711  & 40,000  \\
                             & Test  & 3,813,120  & 40,000  \\
                             & \textbf{Total} & \textbf{38,138,338} & \textbf{400,000} \\ \midrule
\multirow{4}{*}{128 -- 256}  & Train & 61,175,314 & 320,000 \\
                             & Val   & 7,645,806  & 40,000  \\
                             & Test  & 7,644,977  & 40,000  \\
                             & \textbf{Total} & \textbf{76,466,097} & \textbf{400,000} \\ \midrule
\multirow{4}{*}{256 -- 512}  & Train & 122,436,827 & 320,000 \\
                             & Val   & 15,301,762  & 40,000  \\
                             & Test  & 15,303,893  & 40,000  \\
                             & \textbf{Total} & \textbf{153,042,482} & \textbf{400,000} \\ \midrule
\multirow{4}{*}{512 -- 1,024} & Train & 244,714,180 & 320,000 \\
                             & Val   & 30,584,627  & 40,000  \\
                             & Test  & 30,586,768  & 40,000  \\
                             & \textbf{Total} & \textbf{305,885,575} & \textbf{400,000} \\ \midrule
\multirow{4}{*}{1,024 -- 2,048} & Train & 464,065,287 & 303,724 \\
                             & Val   & 58,032,940  & 37,964  \\
                             & Test  & 58,040,040  & 37,968  \\
                             & \textbf{Total} & \textbf{580,138,267} & \textbf{379,656} \\ \midrule
\multirow{4}{*}{2,048 -- 4,096} & Train & 361,195,148 & 118,272 \\
                             & Val   & 45,147,828  & 14,784  \\
                             & Test  & 45,127,513  & 14,784  \\
                             & \textbf{Total} & \textbf{451,470,489} & \textbf{147,840} \\ \midrule
\multirow{4}{*}{4,096 -- 8,192} & Train & 246,548,120 & 40,392  \\
                             & Val   & 30,841,105  & 5,048   \\
                             & Test  & 30,832,237  & 5,052   \\
                             & \textbf{Total} & \textbf{308,221,462} & \textbf{50,492} \\ \bottomrule
\end{tabular}
\end{table}

\section{Training Details}
\label{sec:appendix_training_details}

\subsection{Hyperparameter Settings}
\label{subsec:hyperparams}
The specific hyperparameter configurations for our model architectures and training environment are summarized here. We conduct the training using Low-Rank Adaptation (LoRA) to ensure parameter efficiency. Specifically, we set the LoRA rank to r=16 and the scaling factor to $\alpha$=32, incorporating a dropout rate of 0.05 to mitigate overfitting. The optimization is performed with a learning rate of 0.0001 and a total effective batch size of 128.

\subsection{Training Loss Curves Across Three Stages}
\label{subsec:loss_curves}

To evaluate the optimization stability and convergence of DRIFT, we illustrate the training loss trajectories across its three sequential stages below.

\subsubsection{Stage 1: Latent Fact Reconstruction Pretraining (LFRP)}
\label{subsubsec:loss_lfrp}
\begin{figure}[htbp]
    \centering
    \includegraphics[width=0.95\linewidth]{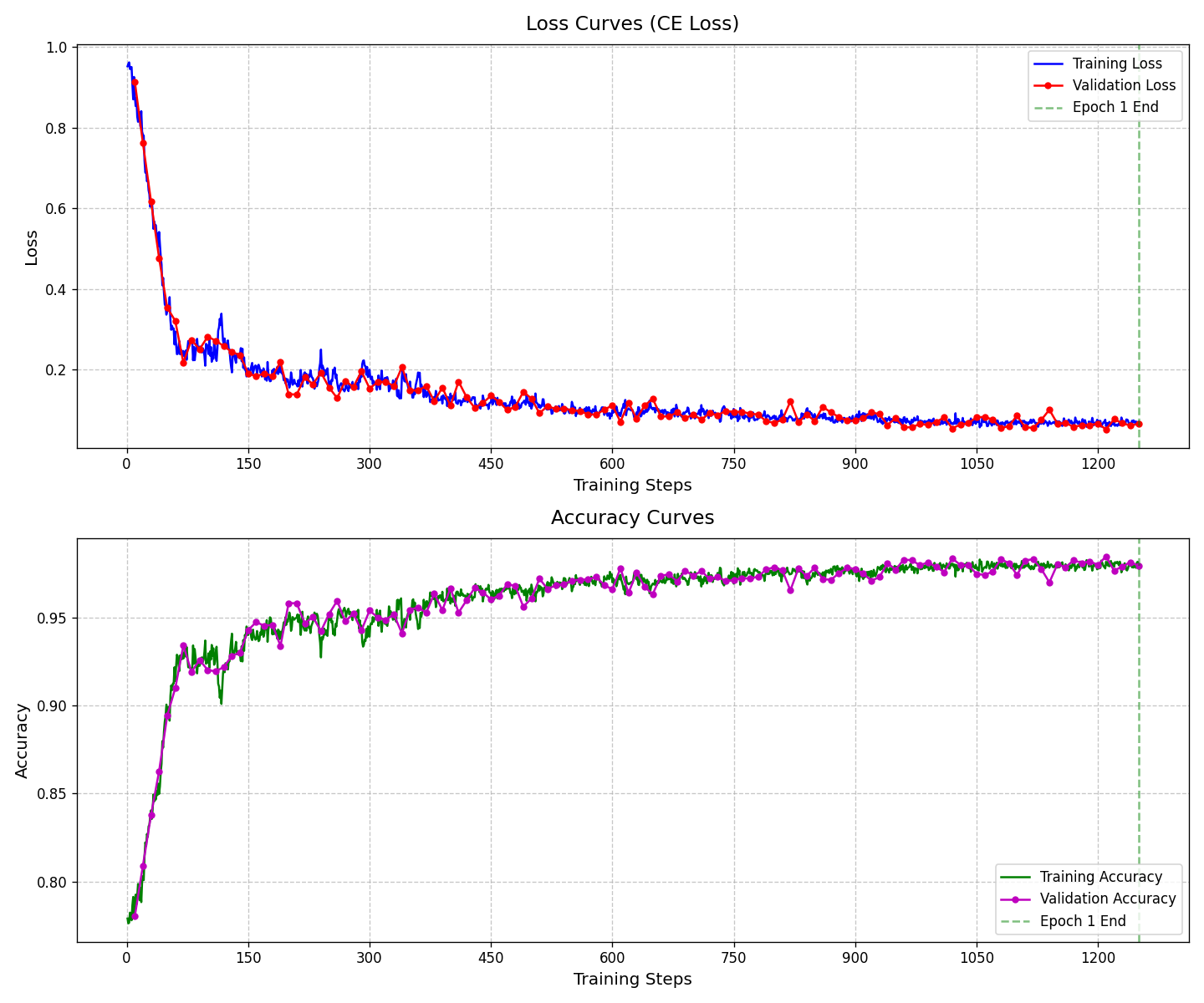}
    \caption{Training loss trajectory of Stage 1}
    \label{fig:stage1_loss}
\end{figure}

The training loss for the LFRP stage is presented in Figure \ref{fig:stage1_loss}. The curve shows a steady decline and eventual plateau, indicating that the knowledge model successfully learned to reconstruct factual content into the latent space with high fidelity.

\subsubsection{Stage 2: Query-Aware Fine-Tuning (QAFT) with Single-Context Dynamic Compression}
\label{subsubsec:loss_qaft_compression}
During this stage, we fine-tune the model on the Dynamic Compression Task. The training objective focuses on the model's ability to compress and project query-relevant knowledge into the reasoning model's embedding space. The loss reflects the efficiency of information bottlenecking under query guidance.
\begin{figure}[htbp]
    \centering
    \includegraphics[width=0.95\linewidth]{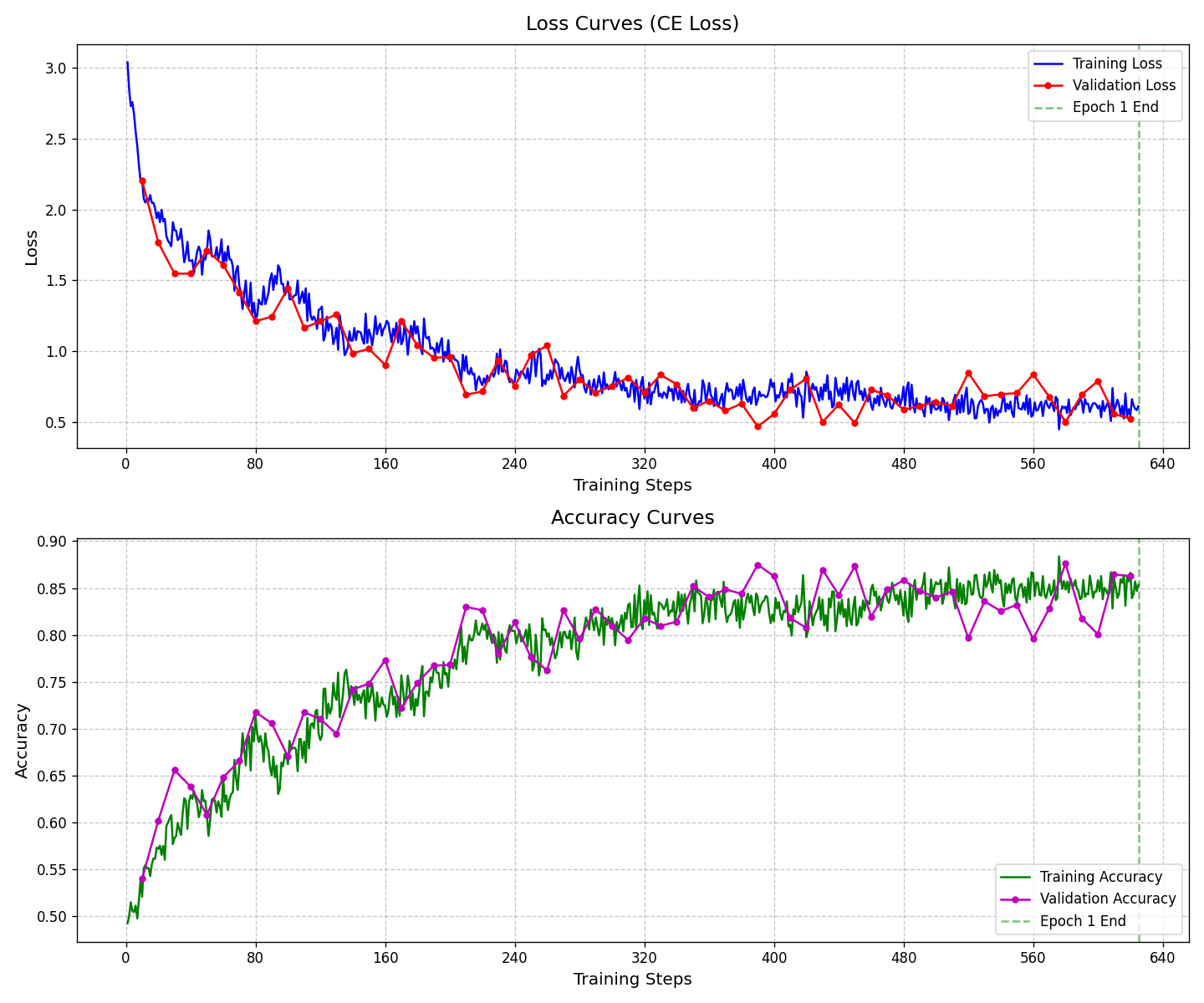}
    \caption{Training loss trajectory of Stage 2}
    \label{fig:stage2_loss}
\end{figure}

\subsubsection{Stage 3: Query-Aware Fine-Tuning (QAFT) with Single-Context Question Answering}
\label{subsubsec:loss_qaft_qa}
The final stage involves end-to-end optimization for the Question-Answering (QA) task. We report the cross-entropy loss during this phase, which demonstrates how the reasoning model effectively utilizes the distilled latent facts to generate accurate and context-grounded answers.
\begin{figure}[htbp]
    \centering
    \includegraphics[width=0.95\linewidth]{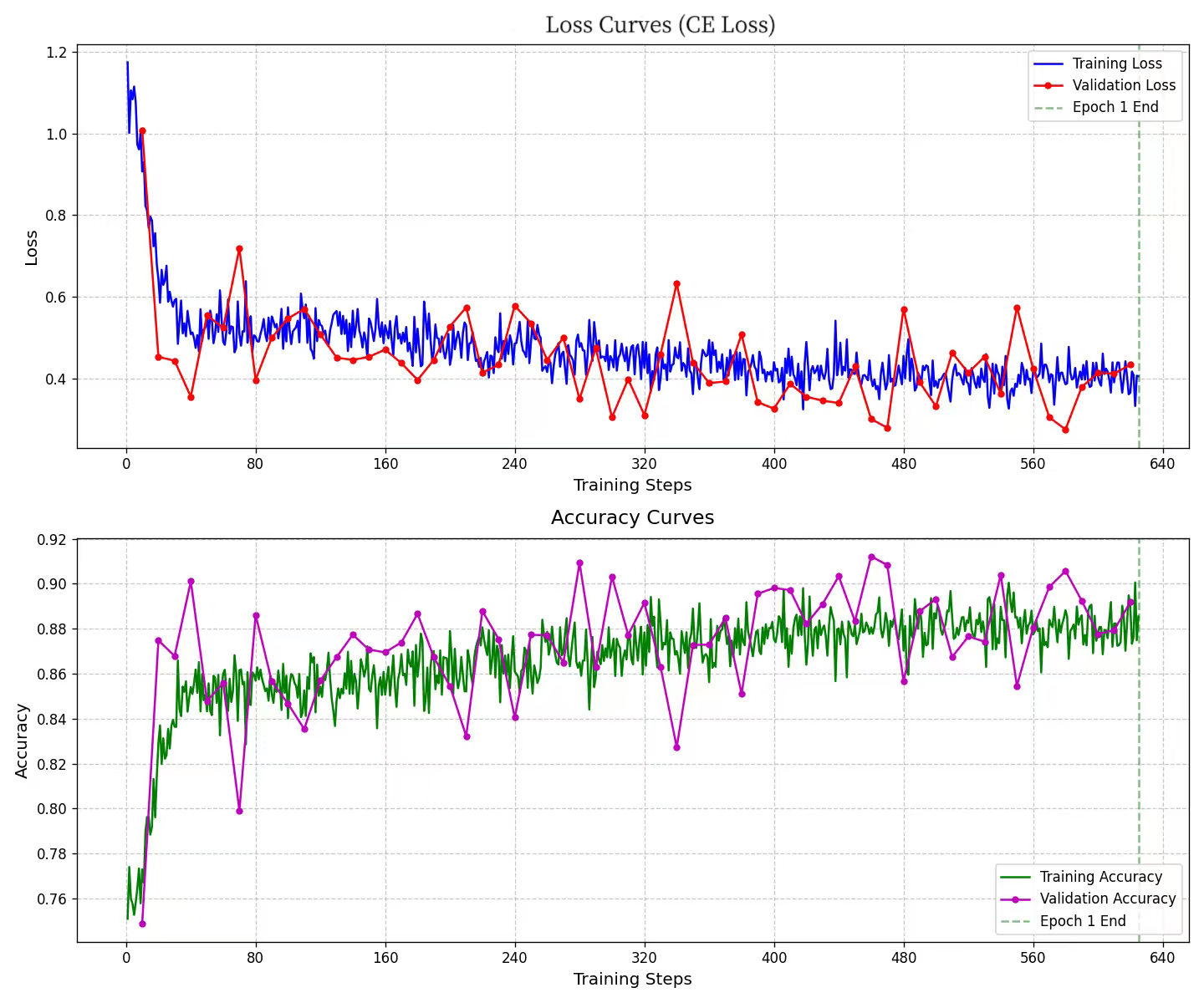}
    \caption{Training loss trajectory of Stage 3}
    \label{fig:stage3_loss}
\end{figure}

\section{Additional Experimental Results}
\label{sec:appendix_additional_exp}

\textbf{RQ5: Does DRIFT outperform the reasoning model with direct input across varied context lengths?}

To compare the performance of DRIFT against the reasoning model with direct input across various context lengths, we evaluated a sampled subset from the BABILong \citep{babilong} benchmark. The resulting comparison is illustrated in the bar chart below:

\begin{figure}[htbp]
    \centering
    \includegraphics[width=1\linewidth]{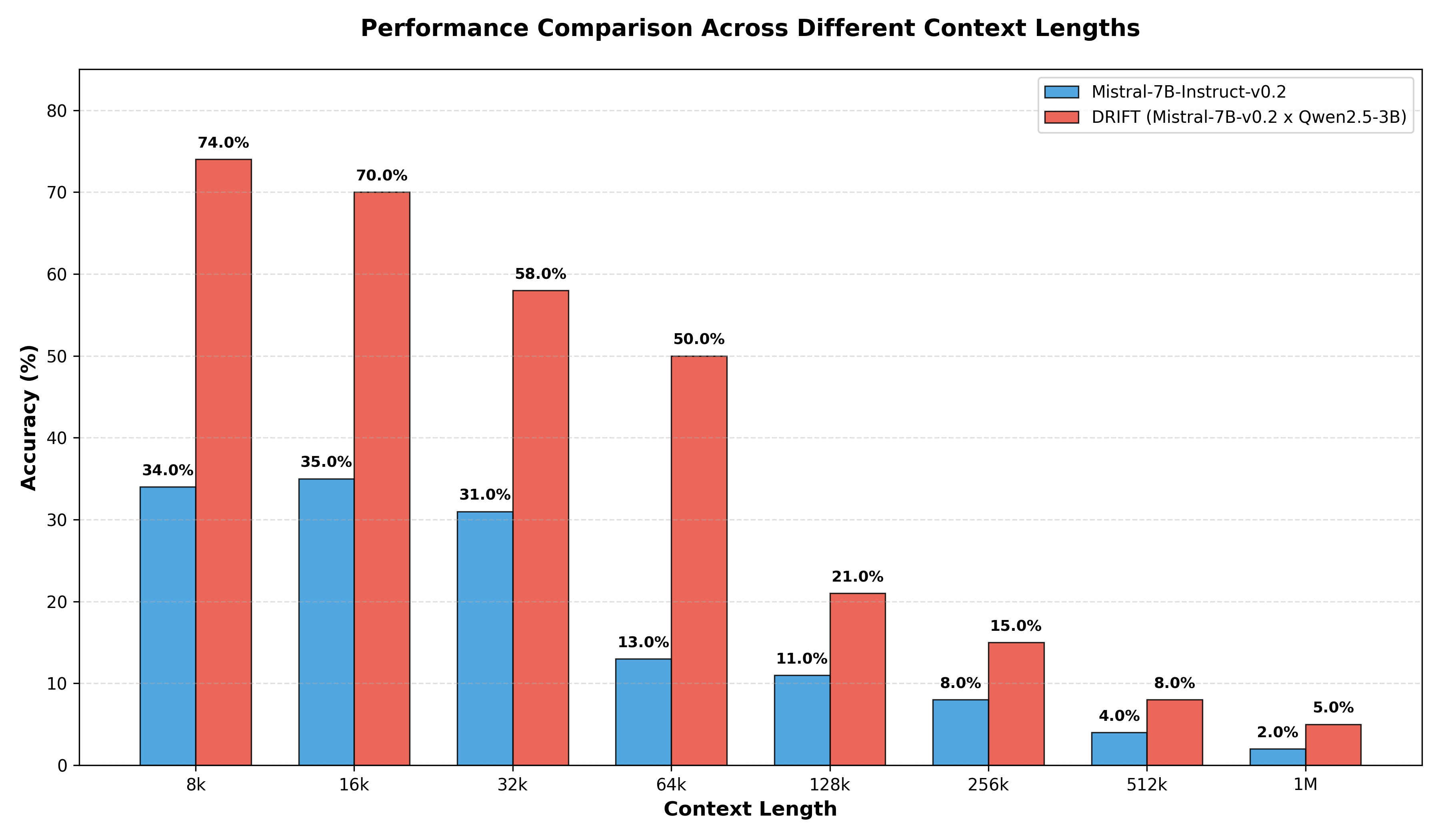}
    \caption{Comparison of accuracy between Mistral-7B-Instruct-v0.2 and DRIFT on a BABILong subset across context lengths from 8k to 1M tokens.}
    \label{fig:bar_chart}
\end{figure}

Experimental results indicate that DRIFT consistently outperforms the vanilla reasoning model across all tested scales, maintaining a substantial performance margin even as the context length extends to 1M~tokens. Notably, at the 64k~context length, DRIFT achieves nearly a $4\times$ accuracy improvement (50.0\% vs. 13.0\%) compared to the direct inference approach.

\textbf{RQ6: Does fine-tuning degrade the reasoning model’s general-purpose capabilities?}

To assess whether the reasoning-oriented fine-tuning (QAFT-QA) affects the model's broad utility, we compare the general-purpose capabilities of the reasoning model before and after our proposed training. The results indicate that the degradation in generic competencies remains minimal. This suggests that our fine-tuning strategy effectively enhances reasoning performance while preserving overall versatility.

\paragraph{Logical Reasoning} To test this capability, we utilized the BBH (Big-Bench Hard) dataset \cite{bbh}, which consists of 23 challenging tasks from the BIG-bench suite that require sophisticated multi-step reasoning where previous language models often underperformed.

\paragraph{Math Reasoning} To test this capability, we evaluate models on the standard GSM8K test set (1,319 problems) \cite{gsm8k}, which is drawn from a benchmark of over eight thousand high-quality grade school math word problems requiring multi-step arithmetic reasoning and logical deduction.

\paragraph{Scientific Knowledge} To test this capability, we utilized the GPQA Diamond dataset \cite{gpqa}, which is a subset of the Graduate-Level Google-Proof Q\&A benchmark. It contains 198 high-quality questions in biology, physics, and chemistry that have been meticulously vetted by experts to ensure they are exceptionally challenging even for highly skilled non-experts with access to the internet.

\paragraph{Instruction Following} To test this capability, we evaluate on the IFBench benchmark \cite{ifbench}, which measures precise instruction-following generalization using a set of verifiable constraints that models must satisfy in their generated outputs.

\paragraph{Code Generation} To test this capability, we utilized the HumanEval dataset \cite{humaneval}, comprising 164 manually crafted Python programming problems used to evaluate the functional correctness of code synthesized from natural language function docstrings.

\paragraph{Factuality} To test this capability, we utilized the HaluEval dataset \cite{halueval}, a large-scale benchmark designed to assess hallucination levels in large language models by testing their ability to recognize and avoid generating factually incorrect information.

\begin{figure}
    \centering
    \includegraphics[width=0.95\linewidth]{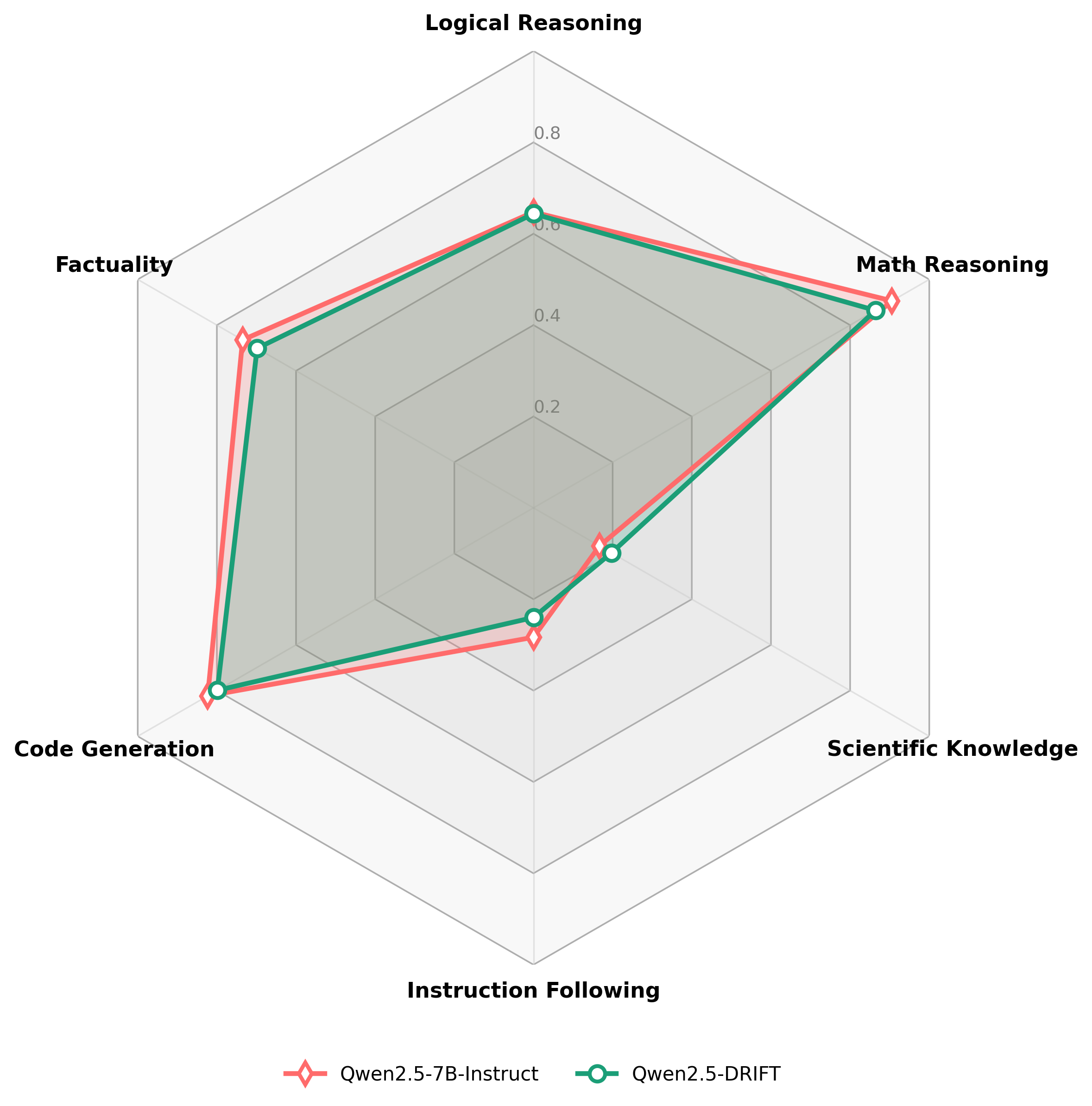}
    \caption{Placeholder radar chart illustrating general-purpose capabilities before and after reasoning-oriented fine-tuning.}
    \label{fig:radar_general}
\end{figure}

\smallskip
\noindent
\textbf{RQ7: Does the DRIFT method remain effective across different combinations of model sizes? What would be the impact of substituting a smaller knowledge model or a larger reasoning model?}
\label{subsec:model_size_scaling}
Experimental analysis of the performance impact when using different size combinations for the knowledge and reasoning models.

\begin{table}[htbp]
\centering
\small
\caption{Performance of different model combinations on LongBenchv2 across different lengths.}
\label{tab:model_combinations}
\begin{tabular}{lcccc}
\toprule
\textbf{Combination} & \multicolumn{4}{c}{\textbf{LongBenchv2}} \\
\cmidrule(lr){2-5}
& \textbf{Short} & \textbf{Medium} & \textbf{Long} & \textbf{Overall} \\
\midrule
Qwen-7B $\times$ 3B   & 36.67 & 28.37 & 33.33 & 32.41 \\
Qwen-7B $\times$ 1.5B  & 27.78 & 29.30 & 37.22 & 31.81 \\
Qwen-14B $\times$ 3B  & 37.03 & 31.63 & 36.11 & 34.39 \\
\bottomrule
\end{tabular}
\end{table}

\smallskip
\noindent
\textbf{RQ8: How does the model align its latent reasoning with explicit evidence during training?}

A critical question in our framework is whether the reasoning model merely replicates explicit textual evidence or instead develops distinct and efficient reasoning strategies within the latent space. This distinction is crucial for determining whether the implicit context functions as a complementary modality rather than a redundant compression.

To examine this behavior, we introduce a diagnostic metric, \textbf{Reasoning Consistency} ($M_{\text{ED}}$), which is monitored during the question-answering task in the \textbf{QAFT stage}. Specifically, $M_{\text{ED}}$ measures the Kullback–Leibler (KL) divergence between the output distributions of the reasoning model when conditioned on compressed latent representations versus explicit textual evidence:
\begin{equation}
\small 
\begin{aligned}
    M_{\mathrm{ED}} = D_{\mathrm{KL}}\Big( & P_{\theta_{\text{rea}}}(\cdot \mid I_{\text{ans}}, E_{\text{dyn}}, \text{E}(Q)) \; \Big\Vert \\
    & P_{\theta_{\text{rea}}}(\cdot \mid I_{\text{ans}}, X_{\text{evi}}, Q) \Big)
\end{aligned}
\end{equation}
Importantly, $M_{\text{ED}}$ is used solely as a non-intrusive diagnostic probe to analyze reasoning behavior and does not participate in gradient backpropagation.

\begin{figure}[h]
    \centering
    \includegraphics[width=0.95\linewidth]{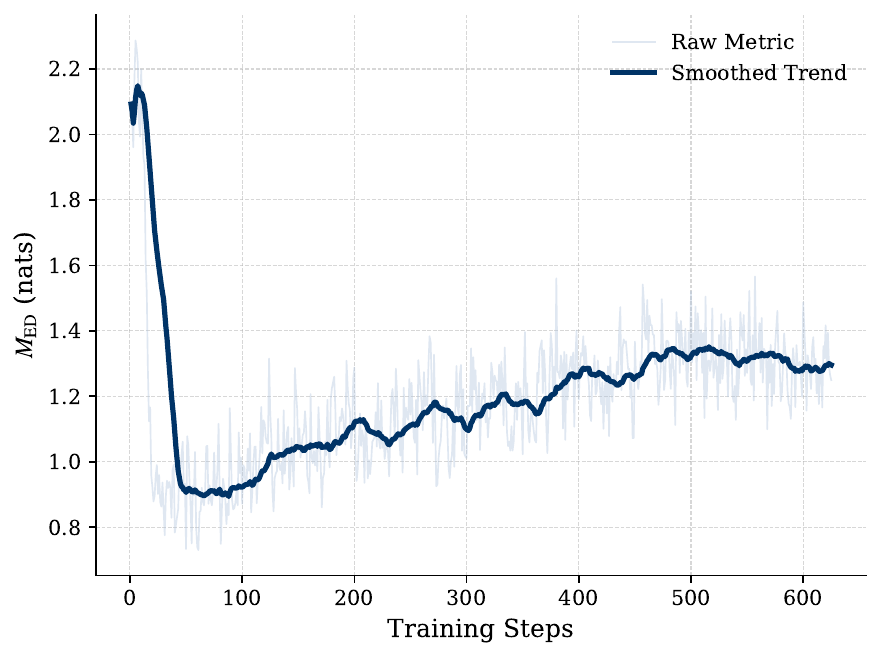}
    \caption{Evolution of the reasoning consistency metric $M_{\text{ED}}$ during traiwoqning. }
    \label{fig:med}
\end{figure}

The trajectory of $M_{\text{ED}}$ in Figure~\ref{fig:med} exhibits a clear \textit{Grounding-then-Specializing} learning pattern with two phases.
In \textbf{Stage I (Rapid Grounding)}, $M_{\text{ED}}$ decreases sharply, indicating that the reasoning model initially aligns its latent representations closely with explicit textual evidence to ensure faithful reasoning.
In \textbf{Stage II (Emergent Specialization)}, $M_{\text{ED}}$ gradually increases and stabilizes, suggesting that the model moves beyond strict textual alignment and develops more specialized and efficient reasoning strategies in the latent space.
Overall, this behavior shows that the implicit context is not a mere compressed copy of the input, but a complementary modality that supports task-oriented reasoning beyond surface-level text.

\section{Other Details}
\subsection{Instructions for DRIFT}
\begin{promptbox}
    $\boldsymbol{I_{sta}}$ ``Given a text passage, condense its core concepts into a set of words. The number of these compressed words is \{num\}. The placeholder of compressed word is `\{COMPRESSION\_TOKEN\}`. The text you need to condense is: <context>{context}</context>  The compressed words are: ``
\end{promptbox}

\begin{promptbox}
    $\boldsymbol{I_{rec}}$ ``Background: <background> \{compressed\_information\} </background>. Please restate the background information above in your own words to convey the same meaning:  ``
\end{promptbox}

\begin{promptbox}
    $\boldsymbol{I_{rec}}$ ``Given several documents and a question, you need to extract the information from the Documents that is relevant to the Question, and condense the core concepts of this knowledge into a set of words. Please note that you are only responsible for extracting information relevant to answering the question. You are not required to reason out the answer yourself. You are not allowed to fabricate information. You may only extract and compress relevant information contained in the documents. Please ensure the completeness and understandability of the compressed knowledge. The number of these compressed words is \{num\}." + f"The placeholder of compressed word is `\{COMPRESSION\_TOKEN\}` The documents are: <Documents> \{document\}</Documents> The question is: <Question>\{question\}</Question>The compressed words of useful information are: .  ``
\end{promptbox}

\begin{promptbox}
    $\boldsymbol{I_{ans}}$ ``You will be provided with a background consisting of \{num\} different paragraphs. Background: <background> \{compressed\_information\} </background>. Please answer the following question based on the background. <Question>\{question\}</Question>\{answer\_prefix\}  ``
\end{promptbox}

\subsection{Text Reconstruction Case}
\begin{figure*}
    \centering
    \includegraphics[width=1\linewidth]{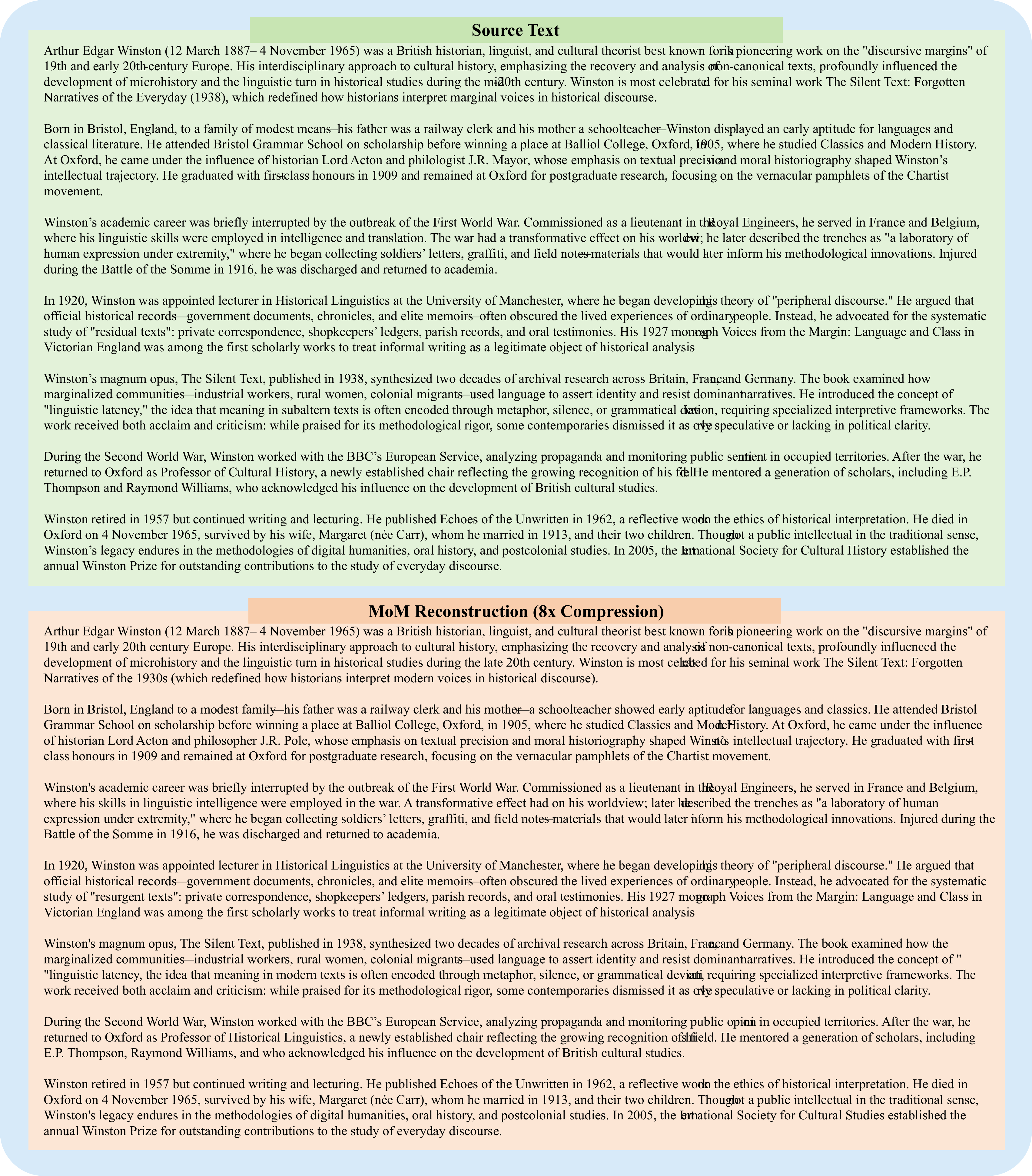}
    \caption{Illustration of the source text alongside its reconstructed counterpart generated by the MoM model.}
    \label{fig:rebuild_compare}
\end{figure*}

A representative case of the source text and the corresponding reconstruction is illustrated in Figure \ref{fig:rebuild_compare}, demonstrating the model's ability to preserve semantic coherence from highly compressed representations.

\end{document}